\documentclass[10pt,twocolumn,letterpaper]{article}
\usepackage[english]{babel} % Explicit language

\usepackage{cvpr}
\usepackage{times}
\usepackage{epsfig}
\usepackage{graphicx}
\usepackage{amsmath, amsthm, amssymb, bm}
\usepackage{booktabs}
\usepackage{multirow}
\usepackage{subfig}
\usepackage[pagebackref=true,breaklinks=true,letterpaper=true,colorlinks,bookmarks=false]{hyperref}
\usepackage{xcolor}

\definecolor{Gray}{gray}{0.95}
\usepackage{colortbl}
\newcolumntype{a}{>{\columncolor{Gray}}c}

\cvprfinalcopy % *** Uncomment this line for the final submission

    \def\cvprPaperID{10766} % *** Enter the CVPR Paper ID here

\ifcvprfinal\fi

\newcommand{\w}{\bm{w}} % A vector-valued random variable.
\begin{document}
%%%%%%%%% TITLE
%\title{A meta-learning free approach for few-shot segmentation}

\title{Few-Shot Segmentation Without Meta-Learning: \\A Good Transductive Inference Is All You Need?}

\author{
Malik Boudiaf \thanks{Corresponding author: malik.boudiaf.1@etsmtl.net} \\
\'ETS Montreal \\
\and
Hoel Kervadec \\
\'ETS Montreal \\
\and
Ziko Imtiaz Masud \\
\'ETS Montreal \\
\and
Pablo Piantanida \\
CentraleSup\'elec-CNRS \\ Universit\'e Paris-Saclay
\and
Ismail Ben Ayed \\
\'ETS Montreal \\
\and
Jose Dolz\\
\'ETS Montreal
}

\maketitle

\begin{abstract}
We show that the way inference is performed in few-shot segmentation tasks has a substantial effect on performances---an aspect often overlooked in the literature in favor of the meta-learning paradigm. We introduce a transductive inference for a given query image, leveraging the statistics of its unlabeled pixels, by optimizing a new loss containing three complementary terms: i) the cross-entropy on the labeled support pixels; ii) the Shannon entropy of the posteriors on the unlabeled query-image pixels; and iii) a global KL-divergence regularizer based on the proportion of the predicted foreground. As our inference uses a simple linear classifier of the extracted features, its computational load is comparable to inductive inference and can be used on top of any base training. Foregoing episodic training and using only standard cross-entropy training on the base classes, our inference yields competitive performances on standard benchmarks in the 1-shot scenarios. As the number of available shots increases, the gap in performances widens: on PASCAL-5$^i$, our method brings about 5\% and 6\% improvements over the state-of-the-art, in the 5- and 10-shot scenarios, respectively. Furthermore, we introduce a new setting that includes domain shifts, where the base and novel classes are drawn from different datasets. Our method achieves the best performances in this more realistic setting. Our code is freely available online: \url{https://github.com/mboudiaf/RePRI-for-Few-Shot-Segmentation}.
\end{abstract}

\section{Introduction}
        Few-shot learning, which aims at classifying instances from unseen classes given only a handful of training examples, has witnessed a rapid progress in the recent years. To quickly adapt to novel classes, there has been a substantial focus on the meta-learning (or learning-to-learn) paradigm \cite{ren2018meta,snell2017prototypical,vinyals2016matching}. Meta-learning approaches popularized the need of structuring the training data into {\em episodes}, thereby simulating the tasks 
        that will be presented at inference. Nevertheless, despite the achieved improvements, several recent image classification works \cite{boudiaf2020transductive,chen2018closer,dhillon2019baseline,GuoECCV20-DS,tian2020rethinking,zikolaplacian} observed that meta-learning might have limited generalization capacity beyond the standard 1- or 5-shot classification benchmarks. For instance, in more realistic settings with domain shifts, simple classification baselines may outperform much more complex meta-learning methods \cite{chen2018closer,GuoECCV20-DS}.

        Deep-learning based semantic segmentation has been generally nurtured from the methodological advances in image classification. Few-shot segmentation, which has gained popularity recently \cite{gairola2020simpropnet,wei2019fss,liu2020crnet,nguyen2019feature,rakelly2018conditional,pfenet,wangfew,wang2019panet,rpmm,yang2020new,zhang2019canet,zhang2020sg}, is no exception. In this setting, a deep segmentation model is first pre-trained on {\em base} classes. Then, model generalization is assessed over few-shot {\em tasks} and novel classes unseen during base training. Each task includes an unlabeled test image, referred to as the {\em query}, along with a few labeled images (the {\em support} set). The recent literature in few-shot segmentation follows the learning-to-learn paradigm, and substantial research efforts focused on the design of specialized architectures and episodic-training schemes for base training. However, i) episodic training itself implicitly assumes that testing tasks have a structure (e.g., the number of support shots) similar to the tasks used at the meta-training stage; 
        and ii) base and novel classes are often assumed to be sampled from the same dataset.

        In practice, those assumptions may limit the applicability of the existing few-shot segmentation methods in realistic scenarios \cite{cao2019theoretical,chen2018closer}. In fact, our experiments proved consistent with findings in few-shot classification when going beyond the standard settings and benchmarks. Particularly, we observed among state-of-the-art methods a saturation in performances \cite{cao2019theoretical} when increasing the number of labeled samples (See Table \ref{tab:n_shots_aggreg}). Also, in line with very recent observations in image classification \cite{chen2018closer}, existing meta-learning methods prove less competitive in cross-domains scenarios (See Table \ref{tab:COCO2PASCAL_results_partial}).
        This casts doubts as to the viability of the current few-shot segmentation benchmarks and datasets; and motivates re-considering the relevance of the meta-learning paradigm, which has become the \textit{de facto} choice in the few-shot segmentation litterature.

        \subsection*{Contributions}
                In this work, we forego meta-learning, and re-consider a simple cross-entropy supervision during training on the base classes for feature extraction. Additionally, we propose a \textit{transductive} inference that better leverages the support-set supervision than the existing methods. Our contributions can be summarized as follows:

                \begin{itemize}
                        \item We present a new transductive inference--\textit{RePRI} (Region Proportion Regularized Inference)--for a given few-shot segmentation task. RePRI optimizes a loss integrating three complementary terms: \textit{i)} a standard cross-entropy on the labeled pixels of the support images; \textit{ii)} the entropy of the posteriors on the query pixels of the test image; and \textit{iii)} a global KL divergence regularizer based on the proportion of the predicted foreground pixels within the test image.
                        RePRI can be used on top of any trained feature extractor, and uses exactly the same information as standard inductive methods for a given few-shot segmentation task.
                        % (i.e., one query image and a few labeled support images), without any additional unlabeled data.
                        % Our transductive inference is based on a simple linear classifier of the extracted features, has a computational load comparable to inductive inference and is modular: it can be used on top of any trained feature extractor. %We call our inference RePRI (Region Proportion Regularized Inference).

                        \item Although we use a basic cross-entropy training on the base classes, without complex meta-learning schemes, RePRI yields highly competitive performances on the standard few-shot segmentation benchmarks, PASCAL-5$^i$ and COCO-20$^i$, with gains around $5\%$ and $6\%$ over the state-of-the-art in the 5- and 10-shot scenarios, respectively.
                        % , while being on par with it in the 1-shot setting. This gap consistently widens as the number of support samples increases, reaching up to $6\%$ in the 10-shot scenario.
                        %This suggests that our transductive inference leverages more effectively the information from the labeled support set of a task.

                        \item We introduce a more realistic setting where, in addition to the usual shift on classes between training and testing data distributions, a shift on the images' feature distribution is also introduced. Our method achieves the best performances in this scenario.

                        \item We demonstrate that a precise region-proportion information on the query object improves substantially the results, with an average gain of 13\% on both datasets. While assuming the availability of such information is not realistic, we show that inexact estimates can still lead to drastic improvements, opening a very promising direction for future research.
                \end{itemize}

\section{Related Work}
        \paragraph{Few-Shot Learning for classification} Meta-learning has become the \textit{de facto} solution to learn novel tasks from a few labeled samples. Even though the idea is not new \cite{schmidhuber1987evolutionary}, it has been revived recently by several popular works in few-shot classification \cite{finn2017model,ravi2016optimization,ren2018meta,snell2017prototypical,vinyals2016matching}. These works can be categorized into gradient- or metric-learning-based methods. Gradient approaches resort to stochastic gradient descent (SGD) to learn the commonalities among different tasks \cite{ravi2016optimization,finn2017model}.
        Metric-learning approaches \cite{vinyals2016matching,snell2017prototypical} adopt deep networks as feature-embedding functions, and compare the distances between the embeddings.
        Furthermore, in a recent line of works, the transductive setting has been investigated for few-shot classification \cite{dhillon2019baseline, boudiaf2020transductive, can, kim2019edge, liu2018learning,team, snell2017prototypical, zikolaplacian}, and yielded performance improvements over inductive inference. These results are in line with established facts in classical transductive inference \cite{vapnik1999overview,joachim99,z2004learning}, well-known to outperform its inductive counterpart on small training sets. To a large extent, these transductive classification works follow well-known concepts in semi-supervised learning, such as graph-based label propagation \cite{liu2018learning}, entropy minimization \cite{dhillon2019baseline} or Laplacian regularization \cite{zikolaplacian}. While the entropy is a part of our transductive loss, we show that it is not sufficient for segmentation tasks, typically yielding trivial solutions.

    \vspace{-1em}
        \paragraph{Few-shot segmentation} Segmentation can be viewed as a classification at the pixel level, and recent efforts mostly went into the design of specialized architectures. Typically, the existing methods use a two-branch comparison framework, inspired from the very popular prototypical networks for few-shot classification \cite{snell2017prototypical}. Particularly, the support images are employed to generate class prototypes, which are later used to segment the query images via a prototype-query comparison module. Early frameworks followed a dual-branch architecture, with two independent branches \cite{shaban2017one,dong2018few,rakelly2018conditional}, one generating the prototypes from the support images and the other segmenting the query images with the learned prototypes. More recently, the dual-branch setting has been unified into a single-branch, employing the same embedding function for both the support and query sets \cite{zhang2020sg,siam2019amp,wang2019panet,rpmm,ppnet}. These approaches mainly aim at exploiting better guidance for the segmentation of query images \cite{zhang2020sg,nguyen2019feature,wangfew,zhang2019pyramid}, by learning better class-specific representations \cite{wang2019panet,liu2020crnet,ppnet,rpmm,siam2019amp} or
        iteratively refining these \cite{zhang2019canet}. Graph CNNs have also been employed to establish more robust correspondences between the support and query images, enhancing the learned prototypes \cite{wangfew}. Alternative solutions to learn better class representations include: imprinting the weights for novel classes \cite{siam2019amp}, decomposing the holistic class representation into a set of part-aware prototypes \cite{ppnet} or mixing several prototypes, each corresponding to diverse image regions \cite{rpmm}.

        \begin{figure*}
                \centering
                \includegraphics[width=\textwidth]{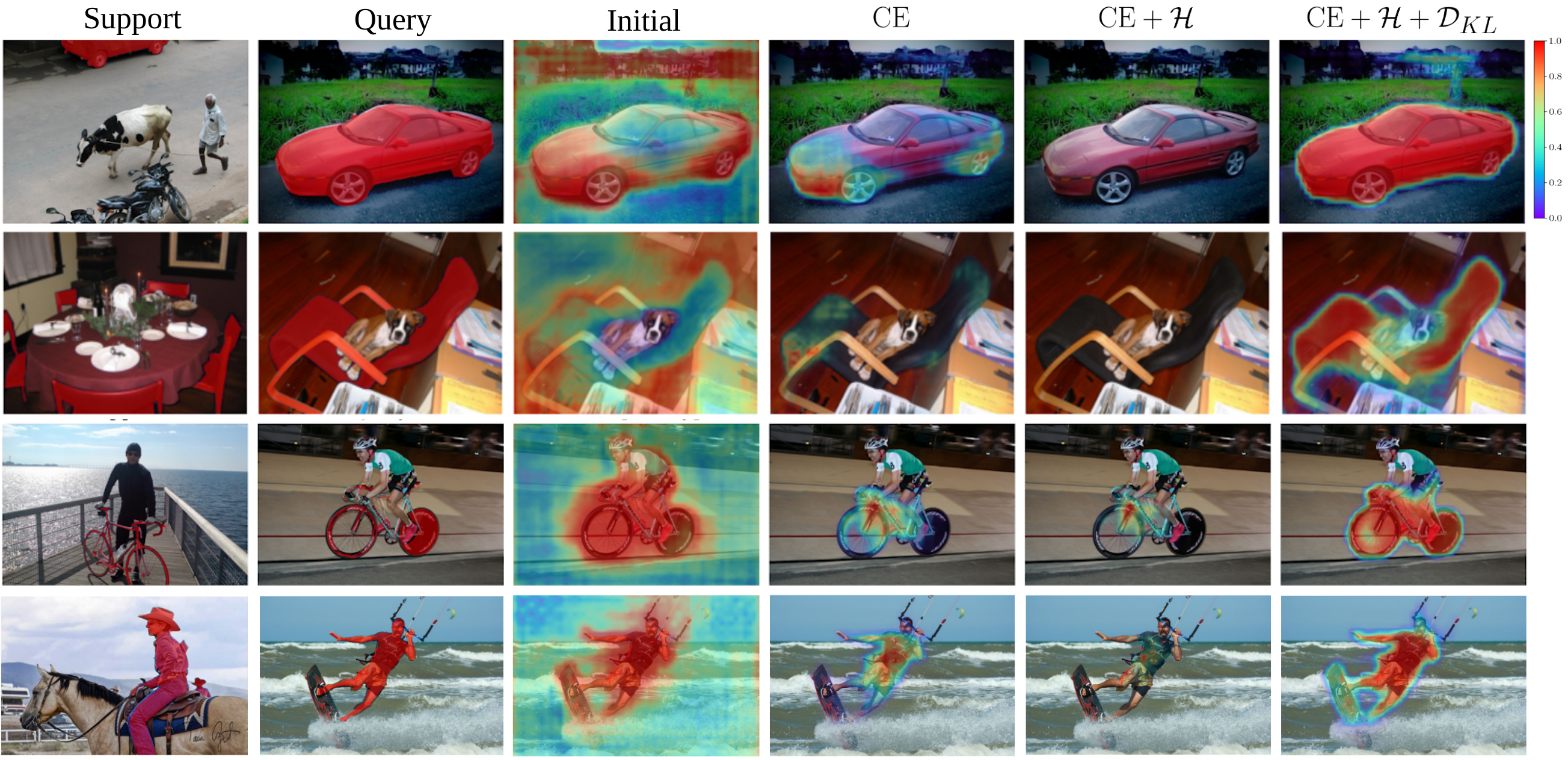}
                \caption{Probability maps for several 1-shot tasks. For each task, the two first columns show the ground truth of support and query. \textit{Initial} column represents the probability map with the initial classifier $\bm \theta^{(0)}$, and the last three columns show the final soft predicted segmentation after finetuning with each of the three losses. Best viewed in colors.}
                \label{fig:loss_comparison}
        \end{figure*}

% \vspace{-1em}
\section{Formulation}
    \subsection{Few-shot Setting}
        Formally, we define a \textit{base} dataset ${\mathcal{D}_{\text{base}}}$ with base semantic classes $\mathcal{Y}_{\text{base}}$, employed for training. Specifically, ${\mathcal{D}_{\text{base}} = \{(x_n, y_n)\}_{n=1}^N}$, $\Omega \subset \mathbb R^{2}$ an image space, ${x_n: \Omega \rightarrow \mathbb R^3}$ an input image, and $y_n: \Omega \rightarrow \{0, 1\}^{|\mathcal{Y}_{base}|}$ its corresponding pixelwise one-hot annotation. At inference, we test our model through a series of $K$-shots tasks. Each $K$-shots task consists of a \textit{support} set $\mathcal S = \{ (x_k, y_k)\}_{k=1}^K$, i.e. $K$ fully annotated images, and one unlabelled query image $x_{\mathcal Q}$, all from the same novel class. This class is randomly sampled from a set of \textit{novel} classes $\mathcal{Y}_{\text{novel}}$ such that $\mathcal{Y}_{\text{base}} \cap \mathcal{Y}_{\text{novel}} = \emptyset$. The goal is to leverage the supervision provided by the support set in order to properly segment the object of interest in the query image.

    \subsection{Base training}
        \paragraph{Inductive bias in episodic training} There exist different ways of leveraging the base set ${\mathcal{D}_{\text{base}}}$. Meta-learning, or \textit{learning to learn}, is the dominant paradigm in the few-shot literature. It emulates the test-time scenario during training by structuring ${\mathcal{D}_{\text{base}}}$ into a series of training tasks. Then, the model is trained on these tasks to learn how to best leverage the supervision from the support set in order to enhance its query segmentation. Recently, Cao et al.  \cite{cao2019theoretical} formally proved that the number of shots $K_{train}$ used in training episodes in the case of prototypical networks represents a learning bias, and that the testing performance saturates quickly when $K_{test}$ differs from $K_{train}$. Empirically, we observed the same trend for current few-shot segmentation methods, with minor improvements from 1-shot to 5-shot performances (Table \ref{tab:PASCAL_results}).

        \vspace{-0.5em}
        \paragraph{Standard training} In practice, the format of the test tasks may be unknown beforehand. Therefore, we want to take as little assumptions as possible on this. This motivates us to employ a feature extractor $f_\phi$ trained with standard cross-entropy supervision on the whole ${\mathcal{D}_{\text{base}}}$ set instead, \textbf{without resorting to episodic training.}

    \subsection{Inference}
        \paragraph{Objective} In what follows, we use $\centerdot$ as a placeholder to denote either a support subscript $k \in \{1, ..., K\}$ or the query subscript $\mathcal{Q}$. At inference, we consider the 1-way segmentation problem: $y_\centerdot: \Omega \rightarrow \{0,1\}^2$ is the function representing the dense \textit{background/foreground} (B/F) mask in image $x_\centerdot$. For both support and query images, we extract features $z_\centerdot:=f_\phi(x_\centerdot)$ and $z_\centerdot: \Psi \rightarrow \mathbb R^C$, where $C$ is the channel dimension in the feature space $\Psi$, with lower pixel resolution $|\Psi| < |\Omega|$.

        Using features $z_\centerdot$, our goal is to learn the parameters $\bm\theta$ of a classifier that properly discriminates foreground from background pixels. Precisely, our classifier $p_\centerdot: \Psi \rightarrow [0, 1]^2$ assigns a (B/F) probability vector to each pixel $j \in \Psi$ in the extracted feature space.

        For each test task, we find the parameters $\bm\theta$ of the classifier by optimizing the following transductive objective:

        \begin{align}\label{eq:main_objective}
            \min_{\bm \theta} \ \text{CE} + \lambda_{\mathcal{H}}~\mathcal{H} + \lambda_{\text{KL}}~ \mathcal{D}_{\text{KL}},
        \end{align}
        where $\lambda_{\mathcal{H}}, \lambda_{\text{KL}} \in \mathbb R$ are non-negative hyper-parameters balancing the effects of the different terms.

        We now describe in details each of the terms in Eq. \eqref{eq:main_objective}:

            \[ \text{CE} = -\frac{1}{K |\Psi|} \sum_{k=1}^K\sum_{j \in \Psi} \widetilde y_k(j)^\top \log(p_k(j))\]
             is the cross-entropy between the downsampled labels $\widetilde y_k$ from support images and our classifier's soft predictions. Simply minimizing this term will often lead to degenerate solutions, especially in the 1-shot setting, as observed in Figure \ref{fig:loss_comparison}---the classifier $\bm \theta$ typically overfits the support set $\mathcal S$, translating into small activated regions on the query image.

            \[ \mathcal H = - \frac{1}{\left| \Psi \right|}\sum_{j \in \Psi} p_{\mathcal Q}(j)^\top \log\left(p_{\mathcal Q}(j)\right)\]
            is the Shannon entropy of the predictions on the query-image pixels. The role of this entropy term is to make the model's predictions more confident on the query image. The use of $\mathcal{H}$ originates from the semi-supervised literature \cite{grandvalet2005semi, miyato2018virtual,berthelot2019mixmatch}. Intuitively, it pushes the decision boundary drawn by the linear classifier towards low-density regions of the extracted query feature space. While this term plays a crucial role in conserving object regions that were initially predicted with only medium confidence, its sole addition to $\text{CE}$ does not solve the problem of degenerate solutions, and may even worsen it in some cases.
           %entropy minimization is widely used \cite{grandvalet2005semi,miyato2018virtual,berthelot2019mixmatch}.

            \[ \mathcal{D}_{\text{KL}} = \widehat p_{\mathcal Q}^{~\top} \log\left(\frac{\widehat p_{\mathcal Q}}{\pi}\right), \]
            with $\widehat p_{\mathcal Q} = \frac{1}{|\Psi|}\sum_{j \in \Psi} p_{\mathcal Q}(j)$,
            is a Kullback-Leibler (KL) Divergence term that encourages the B/F proportion predicted by the model to match a parameter $\pi \in [0, 1]^2$. Notice that the division inside the $\log$ applies element-wise. The joint estimation of parameter $\pi$ in our context is further discussed in a following paragraph. Here, we argue that this term plays a \textbf{key role} in our loss. First, in the case where parameter $\pi$ does not match the exact B/F proportion of the query image, this term still helps avoiding the degenerate solutions stemming from CE and $\mathcal{H}$ minimization. Second, should an accurate estimate of the B/F proportion in the query image be available, it could easily be embedded through this term, resulting in a substantial performance boost, as discussed in Section \ref{sec:experiments}.

        \paragraph{Choice of the classifier} As we  optimize $\bm \theta$ for each task at inference, we want our method to add as little computational load as possible.
        In this regard, we employ a simple linear classifier with learnable parameters $\bm \theta^{(t)}= \{ w^{(t)}, b^{(t)}\}$, with $t$ the current step of the optimization procedure,
        $ w^{(t)} \in \mathbb R^{C}$ the \textit{foreground} prototype and $b^{(t)} \in \mathbb R$ the corresponding bias. Thus, the probabilities $p^{(t)}_k$ and $p^{(t)}_\mathcal Q$ at iteration $t$, for pixel $j \in \Psi$ can be obtained as follow:
        \begin{align}
            p_\centerdot^{(t)}(j) := \begin{pmatrix}
                                        1 - s^{(t)}_\centerdot(j)\\
                                        s^{(t)}_\centerdot(j)
                                        \end{pmatrix},
        \end{align}
        where $s^{(t)}_\centerdot(j) = \text{sigmoid}\left(\tau \left[\cos\left( z_\centerdot(j), w^{(t)} \right)-b^{(t)}\right]\right)$,
        $\tau \in \mathbb R$ is a temperature hyper-parameter and $\cos$ the cosine similarity. The same classifier is used to estimate the support set probabilities $p_k$ and the query predicted probabilities $p_{\mathcal Q}$. At initialization, we set prototype $ w^{(0)}$ to be the average of the foreground support features, i.e. ${ w^{(0)} = \frac{1}{K|\Psi|}\sum_{k=1}^K\sum_{j \in \Psi} {\widetilde y_k(j)_1} z_k(j)}$, with ${\widetilde y_k(j)_1}$ the foreground component of the one-hot label of image $x_k$ at pixel $j$. Initial bias $b^{(0)}$ is set as the mean of the foreground's soft predictions on the query image: $b^{(0)} = \frac{1}{|\Psi|} \sum_{j \in \Psi} p_{\mathcal Q}(j)_1$. Then, $ w^{(t)}$ and $b^{(t)}$ are optimized with gradient descent. The computational footprint of this per-task optimization is discussed in Section \ref{sec:experiments}.

        % \vspace{-11pt}
        \paragraph{Joint estimation of B/F proportion $\pi$} Without additional information, we leverage the model's label-marginal distribution over the query image $\widehat p_{\mathcal Q}^{(t)}$ in order to learn $\pi$ jointly with classifier parameters. Note that minimizing Eq. \eqref{eq:main_objective} with respect to $\pi$ yields ${\pi}^{(t)}={\widehat p}_{\mathcal Q}^{(t)}$. Empirically, we found that, after initialization, updating $\pi$ only once during optimization, at a later iteration, $t_{\pi}$ was enough:
        \begin{equation}\label{eq:prior_init} \pi^{(t)} =
            \begin{cases}
                \widehat{p}_\mathcal{Q}^{(0)} & 0 \leq t \leq t_{\pi}\\[4pt]
                \widehat{p}_\mathcal{Q}^{(t_{\pi})} & t > t_{\pi}.\\
            \end{cases}
        \end{equation}
        Intuitively, the entropy term $\mathcal{H}$ helps gradually refine initially blurry soft predictions (third column in Fig. \ref{fig:loss_comparison}), which turns $\widehat p_{\mathcal Q}^{(t)}$ into an improving estimate of the true B/F proportion. A quantitative study of this phenomenon is provided in Section \ref{sec:ablation}.
        % $\pi^{(t)}$ is then updated once during optimization (more details in Section \ref{sec:experiments}).
        Therefore, our inference can be seen as a joint optimization over $\bm \theta$ and $\pi$, with $\mathcal{D}_{\text{KL}}$ serving as a {\em self-regularization} that prevents the model's marginal distribution $\widehat p_{\mathcal Q}^{(t)}$ from diverging.

        \paragraph{Oracle case with a known $\pi$} As an upper bound, we also investigate the \textit{oracle} case, where we have access to the true B/F proportion in $x_\mathcal{Q}$:
        \begin{equation}
            \pi^* = \frac{1}{|\Psi|} \sum_{j \in \Psi}\widetilde y_{\mathcal Q}(j).
        \end{equation}

\section{Experiments}\label{sec:experiments}

    \paragraph{Datasets} We resort to two public few-shot segmentation benchmarks, PASCAL-5$^i$ and COCO-20$^i$, to evaluate our method. PASCAL-5$^i$ is built from PASCALVOC 2012 \cite{PASCAL_voc}, and contains 20 object categories split into 4 folds. For each fold, 15 classes are used for training and the remaining 5 categories for testing. COCO-20$^i$ is built from MS-COCO \cite{ms-COCO} and is more challenging, as it contains more samples, more classes and more instances per image. Similar to PASCAL-5$^i$, COCO-20$^i$ dataset is divided into 4-folds with 60 base classes and 20 test classes in each fold.

    \paragraph{Training} We build our model based on PSPNet \cite{pspnet} with Resnet-50 and Resnet-101 \cite{resnet} as backbones. We train the feature extractor with standard cross-entropy over the base classes during 100 epochs on PASCAL-5$^i$, and 20 epochs on COCO-20$^i$, with batch size set to 12. We use SGD as optimizer with the initial learning rate set to 2.5e$-3$ and we use cosine decay. Momentum is set to 0.9, and weight decay to 1e$-4$. Label smoothing is used with smoothing parameter $\epsilon=0.1$. We did not use multi-scaling, nor deep supervision, unlike the original PSPNet paper \cite{pspnet}. As for data augmentations, we only use random mirror flipping.

    \begin{table*}[t!!!]
        \centering
        \small
        \caption{Results of 1-way 1-shot and 1-way 5-shot segmentation on PASCAL-5$^i$ using the mean-IoU. Best results in bold.}
        \resizebox{\textwidth}{!}{
            \begin{tabular}{lcccccacccca}
                \toprule
                 & & \multicolumn{5}{c}{1 shot} & \multicolumn{5}{c}{5 shot} \\
                 \cmidrule(lr){3-7}\cmidrule(lr){8-12}
                 Method & Backbone & Fold-0 & Fold-1 & Fold-2 & Fold-3 & Mean & Fold-0 & Fold-1 & Fold-2 & Fold-3 & Mean \\
                 \midrule
                 OSLSM \cite{shaban2017one} (BMVC'18) & \multirow{9}{*}{VGG-16} &33.6 & 55.3 & 40.9 & 33.5 & 40.8 & 35.9 & 58.1 & 42.7 & 39.1& 43.9 \\
                 co-FCN \cite{rakelly2018conditional} (ICLRW'18) &  &36.7 & 50.6 & 44.9 & 32.4 & 41.1 & 37.5 & 50.0 & 44.1 & 33.9 & 41.4  \\
                 AMP \cite{siam2019amp} (ICCV'19) & &41.9 & 50.2 & 46.7 & 34.7 & 43.4 & 41.8 & 55.5 & 50.3 & 39.9 & 46.9   \\
                 PANet \cite{wang2019panet} (ICCV'19) & &42.3 & 58.0 & 51.1 & 41.2 & 48.1 & 51.8 & 64.6 & 59.8 & 46.5 & 55.7  \\
                 FWB \cite{nguyen2019feature} (ICCV'19) & &47.0 & 59.6 & 52.6 & 48.3 & 51.9 & 50.9 & 62.9 & 56.5 & 50.1 & 55.1  \\
                 SG-One \cite{zhang2020sg} (TCYB'20) & & 40.2 & 58.4 & 48.4 & 38.4 & 46.3 & 41.9 & 58.6 & 48.6 & 39.4 & 47.1\\
                 CRNet \cite{liu2020crnet} (CVPR'20)& & - & - & - & - & 55.2 & - & - & - & - & 58.5  \\
                 FSS-1000 \cite{wei2019fss} (CVPR'20)& &- & - & - & - & - & 37.4 & 60.9 & 46.6 & 42.2 & 56.8   \\
                 RPMM \cite{ppnet} (ECCV'20) & & 47.1 & 65.8 & 50.6 & 48.5 & 53.0 & 50.0&  66.5 & 51.9 & 47.6 & 54.0 \\
                 \hline
                 CANet  \cite{zhang2019canet} (CVPR'19) & \multirow{9}{*}{ResNet50} & 52.5 & 65.9 & 51.3 & 51.9 &  55.4 & 55.5 & 67.8 & 51.9 & 53.2  & 57.1 \\
                 PGNet \cite{zhang2019pyramid} (ICCV'19)  &  &  56.0 & 66.9 & 50.6 & 50.4 &  56.0 & 57.7 & 68.7 & 52.9 & 54.6  & 58.5 \\
                 CRNet \cite{liu2020crnet} (CVPR'20) &  & - & - & - & - & 55.7 & - & - & - & - & 58.8  \\
                 SimPropNet  \cite{gairola2020simpropnet} (IJCAI'20) &  & 54.9 & 67.3 & 54.5 & 52.0 & 57.2& 57.2 & 68.5 & 58.4 & 56.1 & 60.0 \\
                 LTM \cite{yang2020new} (MMMM'20) & & 52.8 & 69.6 & 53.2 & 52.3 & 57.0& 57.9 & 69.9 & 56.9 & 57.5 & 60.6 \\

                 RPMM \cite{rpmm} (ECCV'20) & &55.2 & 66.9 & 52.6 & 50.7 & 56.3 & 56.3 & 67.3 &  54.5 & 51.0 & 57.3  \\
                 PPNet \cite{ppnet} (ECCV'20)* & &47.8 & 58.8 & 53.8 & 45.6 & 51.5 & 58.4 & 67.8 &  64.9 &56.7 &  62.0  \\
                 PFENet \cite{pfenet} (TPAMI'20) & & \textbf{61.7} & \textbf{69.5} & 55.4 & \textbf{56.3} & \textbf{60.8} & 63.1 & 70.7 & 55.8 & 57.9 & 61.9 \\
                 RePRI (ours) & &  60.2 & 67.0 & \textbf{61.7} & 47.5 & 59.1 & \textbf{64.5} & \textbf{70.8} & \textbf{71.7} & \textbf{60.3} & \textbf{66.8} \\
                \hline
                 Oracle-RePRI  & ResNet50 & 72.4 & 78.0 & 77.1 & 65.8 & 73.3 & 75.1 & 80.8 & 81.4 & 74.4 & 77.9 \\
                 \hline
                 FWB \cite{nguyen2019feature} (ICCV'19)& \multirow{4}{*}{ResNet101} &51.3 & 64.5 & 56.7 & 52.2 & 56.2 & 54.9 & 67.4 &  62.2 & 55.3 & 59.9  \\
                 DAN \cite{wangfew} (ECCV'20)&  & 54.7 &  68.6 & 57.8 & 51.6 & 58.2 & 57.9 &  69.0 & 60.1 & 54.9 & 60.5  \\
                 PFENet \cite{pfenet} (TPAMI'20)   &  & \textbf{60.5} & \textbf{69.4} & 54.4 & \textbf{55.9} & \textbf{60.1} & 62.8 & 70.4 & 54.9 & 57.6 & 61.4 \\
                 RePRI (ours) & &  59.6 & 68.6 & \textbf{62.2} & 47.2 & 59.4 & \textbf{66.2} & \textbf{71.4} & \textbf{67.0} & \textbf{57.7} & \textbf{65.6} \\
                 \hline
                 Oracle-RePRI  & ResNet101 & 73.9 & 79.7 & 76.1 & 65.1 & 73.7 & 76.8 & 81.7 & 79.5 & 74.5 & 78.1 \\
                 \bottomrule
                \multicolumn{8}{l}{\scriptsize{* We report the results where no additional unlabeled data is employed.}}\\
            \end{tabular}
        }
        \label{tab:PASCAL_results}
    \end{table*}

    \paragraph{Inference} At inference, following previous works \cite{ppnet, wang2019panet}, all images are resized to a fixed $417 \times 417$ resolution. For each task, the classifier $\bm \theta$ is built on top of the features from the penultimate layer of the trained network. For our model with ResNet-50 as backbone, this results in a $53\ \times 53 \times 512$ feature map. SGD optimizer is used to train $\bm \theta$, with a learning rate of 0.025. For each task, a total of 50 iterations are performed. The parameter $t_{\pi}$ is set to 10. For the main method, the weights $\lambda_{\mathcal{H}}$ and $\lambda_{\text{KL}}$ are both initially set to $1/K$, such that the CE term plays a more important role as the number of shots $K$ grows. For $t \geq t_{\pi}$, $\lambda_{\text{KL}}$ is increased by 1 to further encourage the predicted proportion close to $\pi^{(t_\pi)}$. Finally, the temperature $\tau$ is set to 20.

    \paragraph{Evaluation} We employ the widely adopted mean Intersection over Union (mIoU). Specifically, for each class, the classwise-IoU is computed as the sum over all samples within the class of the intersection over the sum of all unions. Then, the mIoU is computed as the average over all classes of the classwise-IoU. Following previous works \cite{ppnet}, 5 runs of 1000 tasks each are computed for each fold, and the average mIoU over runs is reported.

    \subsection{Benchmark results}
        \paragraph{Main method} First, we investigate the performance of the proposed method in the popular 1-shot and 5-shot settings on both PASCAL-5$^i$ and COCO-20$^i$, whose results are reported in Table \ref{tab:PASCAL_results} and \ref{tab:COCO_results}. Overall, we found that our method compares competitively with state-of-the-art approaches in the 1-shot setting, and significantly outperforms recent methods in the 5-shot scenario. Additional qualitative results on PASCAL-5$^i$ are shown in the supplemental material.

        \begin{table*}[t!!]
            \centering
            \small
            \caption{Results of 1-way 1-shot and 1-way 5-shot segmentation on COCO-20$^i$ using mean-IoU metric. Best results in bold.}
            \resizebox{\textwidth}{!}{
                \begin{tabular}{lcccccacccca}
                    \toprule
                     & & \multicolumn{5}{c}{1 shot} & \multicolumn{5}{c}{5 shot} \\
                     \cmidrule(lr){3-7}\cmidrule(lr){8-12}
                     Method & Backbone & Fold-0 & Fold-1 & Fold-2 & Fold-3 & Mean & Fold-0 & Fold-1 & Fold-2 & Fold-3 & Mean \\
                     \midrule
                     PPNet* \cite{ppnet} (ECCV'20) & \multirow{4}{*}{ResNet50} & 34.5 & 25.4 & 24.3 & 18.6 & 25.7 & 48.3 & 30.9 & 35.7 & 30.2 & 36.2\\

                     RPMM \cite{rpmm} (ECCV'20)  &  & 29.5 & 36.8 & 29.0 & 27.0 & 30.6 & 33.8 & 42.0 & 33.0 & 33.3 &  35.5\\
                    %  RePRI (ours) & Resnet-50 & 34.1 & 36.2 & 29.9 & 32.9 & 33.2 & 44.2 & 42.3 & 37.8 & 41.2 & 41.3 \\
                     PFENet \cite{pfenet} (TPAMI'20) &  &  \textbf{36.5} & \textbf{38.6} & \textbf{34.5} & \textbf{33.8} & \textbf{35.8} & 36.5 & 43.3 & 37.8 & 38.4 & 39.0 \\
                     RePRI (ours) &  & 31.2 & 38.1 & 33.3 & 33.0 & 34.0 & \textbf{38.5} & \textbf{46.2} & \textbf{40.0} & \textbf{43.6} & \textbf{42.1} \\
                     \hline
                     Oracle-RePRI & ResNet50 & 49.3 & 51.4 & 38.2 & 41.6 & 45.1 & 51.5 & 60.8 & 54.7 & 55.2 & 55.5 \\
                     \bottomrule
                    \multicolumn{8}{l}{\scriptsize{* We report the results where no additional unlabeled data is employed.}}
                \end{tabular}
            }
            \label{tab:COCO_results}
        \end{table*}

        \vspace{-1em}
        \paragraph{Beyond 5-shots} In the popular learning-to-learn para-digm, the number of shots leveraged during the meta-training stage has a direct impact on the performance at inference \cite{cao2019theoretical}. Particularly, to achieve the best performance, meta-learning based methods typically require the numbers of shots used during meta-training to match those employed during meta-testing. To demonstrate that the proposed method is more robust against differences on the number of labeled support samples between the base and test sets,
        we further investigate the 10-shot scenario. Particularly, we trained the methods in \cite{pfenet,rpmm} by using one labeled sample per class, i.e., 1-shot task, and test the models on a 10-shots task.
        % i.e., different number of shots in training and inference, we introduce a novel setting,         where models are trained with only one-shot and leverage 10-shots for testing.
        Interestingly, we show that the gap between our method and current state-of-the-art becomes larger as the number of support images increases (Table \ref{tab:n_shots_aggreg}), with significant gains of 6\% and 4\% on PASCAL-5$^i$ and COCO-20$^i$, respectively. These results suggest that our transductive inference  leverages more effectively  the information conveyed in the labeled support set of a given task.

        \begin{table}
            \small
            \centering
            \caption{Aggregated results for 1-way 1-, 5- and 10-shot tasks with Resnet50 as backbone and averaged over 4 folds. Per fold results are available in the supplementary material.}
            \resizebox{\columnwidth}{!}{
                \begin{tabular}{lcccccc}
                    \toprule
                     & \multicolumn{3}{c}{PASCAL-5$^i$} & \multicolumn{3}{c}{COCO-20$^i$} \\
                     \cmidrule(lr){2-4}\cmidrule(lr){5-7}
                     Method  & 1-S & 5-S & 10-S & 1-S & 5-S & 10-S \\
                     \midrule
                     RPMM \cite{rpmm}  & 56.3 & 57.3 & 57.6 & 30.6 & 35.5 & 33.1 \\
                     PFENet \cite{pfenet} & \textbf{60.8} & 61.9 & 62.1 & \textbf{35.8} & 39.0 & 39.7 \\
                     RePRI (ours) & 59.1 & \textbf{66.8} & \textbf{68.2} & 34.0 & \textbf{42.1} & \textbf{44.4} \\
                     \hline
                     Oracle-RePRI & 73.3 & 77.9 & 78.6 & 45.1 & 55.5 & 58.7 \\
                     \bottomrule
                \end{tabular}
            }
            \label{tab:n_shots_aggreg}
        \end{table}

        \vspace{-1em}
        \paragraph{Oracle results} We now investigate the ideal scenario where an oracle provides the exact foreground/background proportion in the query image, such that $\pi^{(t)}=\pi^*, \forall t$. Reported results in this scenario, referred to as \textit{Oracle} (Table \ref{tab:PASCAL_results} and \ref{tab:COCO_results}) show impressive improvements over both our current method and all previous works, with a consistent gain across datasets and tasks. Particularly, these values range from 11\% and 14 \% on both PASCAL-5$^i$ and COCO-20$^i$ and in both 1-shot and 5-shot settings. We believe that these findings convey two important messages. First, it proves that there exists a simple linear classifier that can largely outperform state-of-the-art meta-learning models, while being built on top of a feature extractor trained with a standard cross-entropy loss. Second, these results indicate that having a precise size of the query object of interest acts as a strong regularizer. This suggests that more efforts could be directed towards properly constraining the optimization process of $\w$ and $b$, and opens a door to promising avenues.

    \subsection{Domain shift}
        We introduce a more realistic, cross-domain setting (COCO-20$^i$ to PASCAL-VOC).
        We argue that such setting is a step towards a more realistic evaluation of these methods, as it can assess the impact on performances caused by a domain shift between the data training distribution and the testing one. We believe that this scenario can be easily found in practice, as even slight alterations in the data collection process might result in a distributional shift.
        We reproduce the scenario where a large labeled dataset is available (e.g., COCO-20$^i$), but the evaluation is performed on a target dataset with a different feature distribution (e.g., PASCAL-VOC).
        As per the original work \cite{ms-COCO}, significant differences exist between the two original datasets. For instance, images in MS-COCO have on average 7.7 instances of objects coming from 3.5 distinct categories, while PASCAL-VOC only has an average of 3 instances from 2 distinct categories.

        \paragraph{Evaluation}  We reuse models trained on each fold of COCO-20$^i$ and generate tasks using images from all the classes in PASCAL-VOC that were not used during training. For instance, fold-0 of this setting means the model was trained on fold-0 of COCO-20$^i$ and tested on the whole PASCAL-VOC dataset, after removing the classes seen in training.
        A complete summary of all the folds is available in the Supplemental material.

        \paragraph{Results} We reproduced and compared to the two best performing methods \cite{pfenet, ppnet} using their respective official GitHub repositories.
        Table \ref{tab:COCO2PASCAL_results_partial} summarizes the results for the  1-shot and 5-shot cross-domain experiments. We observe that in the presence of domain-shift, our method outperforms existing methods in both 1-shot and 5-shot scenarios, with again the improvement jumping from 2\% in 1-shot to 4\% in 5-shot.

    \begin{table}
        \centering
        \small
        \caption{Aggregated domain-shift results, averaged over 4 folds, on COCO-20$^i$ to PASCAL-VOC. Best results in bold. Per-fold results are available in the supplementary material.}
        \resizebox{0.9\columnwidth}{!}{
            \begin{tabular}{lccc}
                \toprule
                & & \multicolumn{2}{c}{COCO $\rightarrow$ PASCAL} \\
                \cmidrule{3-4}
                 Method & Backbone & 1 shot & 5 shot \\
                 \midrule
                 RPMM \cite{rpmm}  & \multirow{3}{*}{ResNet50} & 49.6 & 53.8 \\
                 PFENet \cite{pfenet} & & 61.1 & 63.4 \\
                 RePRI (ours) & & \textbf{63.2} & \textbf{67.7} \\
                 \hline
                 Oracle-RePRI & Resnet-50 & 76.2 & 79.7 \\
                 \bottomrule
            \end{tabular}
        }
        \label{tab:COCO2PASCAL_results_partial}
    \end{table}

    \subsection{Ablation studies}\label{sec:ablation}

        \begin{table*}
            \centering
            \small
            \caption{Ablation study on the effect of each term in our loss in Eq. (\ref{eq:main_objective}), evaluated on PASCAL-5$^i$.}
            \begin{tabular}{lccccacccca}
                \toprule
                 & \multicolumn{5}{c}{1 shot} & \multicolumn{5}{c}{5 shot} \\
                 \cmidrule(lr){2-6}\cmidrule(lr){7-11}
                 Loss & Fold-0 & Fold-1 & Fold-2 & Fold-3 & Mean & Fold-0 & Fold-1 & Fold-2 & Fold-3 & Mean \\
                 \midrule
                 $\text{CE}$ & 39.7 & 49.3 & 37.3 & 27.5 & 38.5 & 56.5 & 66.4 & 60.1 & 49.0 & 58.0 \\
                 $\text{CE} + \mathcal{H}$ & 45.7 & 61.7 & 48.2 & 36.4 & 48.0 & 56.8 & 68.5 & 61.3 & 47.0 & 58.4 \\
                 $\text{CE} + \mathcal{H} + \mathcal{D}_{\text{KL}}$ & \textbf{60.2} & \textbf{67.0} & \textbf{61.7} & \textbf{47.5} & \textbf{59.1} & \textbf{64.5} & \textbf{70.8} & \textbf{71.7}& \textbf{60.3} & \textbf{66.8} \\
                 \bottomrule
            \end{tabular}
            \label{tab:ablation_loss}
        \end{table*}

        \paragraph{Impact of each term in the main objective} While Fig. \ref{fig:loss_comparison} provides \textit{qualitative} insights on how each term in Eq. \eqref{eq:main_objective} affects the final prediction, this section provides a \textit{quantitative} evaluation of their impact, evaluated on PASCAL-5$^i$ (Table \ref{tab:ablation_loss}). Quantitative results confirm the qualitative insights observed in Fig. \ref{fig:loss_comparison}, as both $\text{CE}$ and $\text{CE} + \mathcal{H}$ losses drastically degrade the performance compared to the proportion-regularized loss, i.e., $\text{CE} + \mathcal{H} + \mathcal{D}_{\text{KL}}$. For example, in the 1-shot scenario, simply minimizing the $\text{CE}$ results in more than 20\% of difference compared to the proposed model. In this case, the prototype $\w$ tends to overfit the support sample and only activates regions of the query object that strongly correlate with the support object. Such behavior hampers the performance when the support and query objects exhibit slights changes in shape or histogram colors, for example, which may be very common in practice. Adding the entropy term $\mathcal{H}$ to $\text{CE}$ partially alleviates this problem, as it tends to reinforce the model in being confident on positive pixels initially classified with mid or low confidence. Nevertheless, despite improving the naive $\text{CE}$ based model, the gap with the proposed model remains considerably large, with 10\% difference. One may notice that the differences between $\text{CE}$ and $\text{CE} + \mathcal{H} + \mathcal{D}_{\text{KL}}$ decrease in the 5-shot setting, since overfitting 5 support samples simultaneously becomes more difficult. The results from this ablation experiment reinforce our initial hypothesis that the proposed KL term based on the size parameter $\pi$ acts as a strong regularizer.

\begin{figure}[h]
            \centering
            \includegraphics[width=0.485\textwidth]{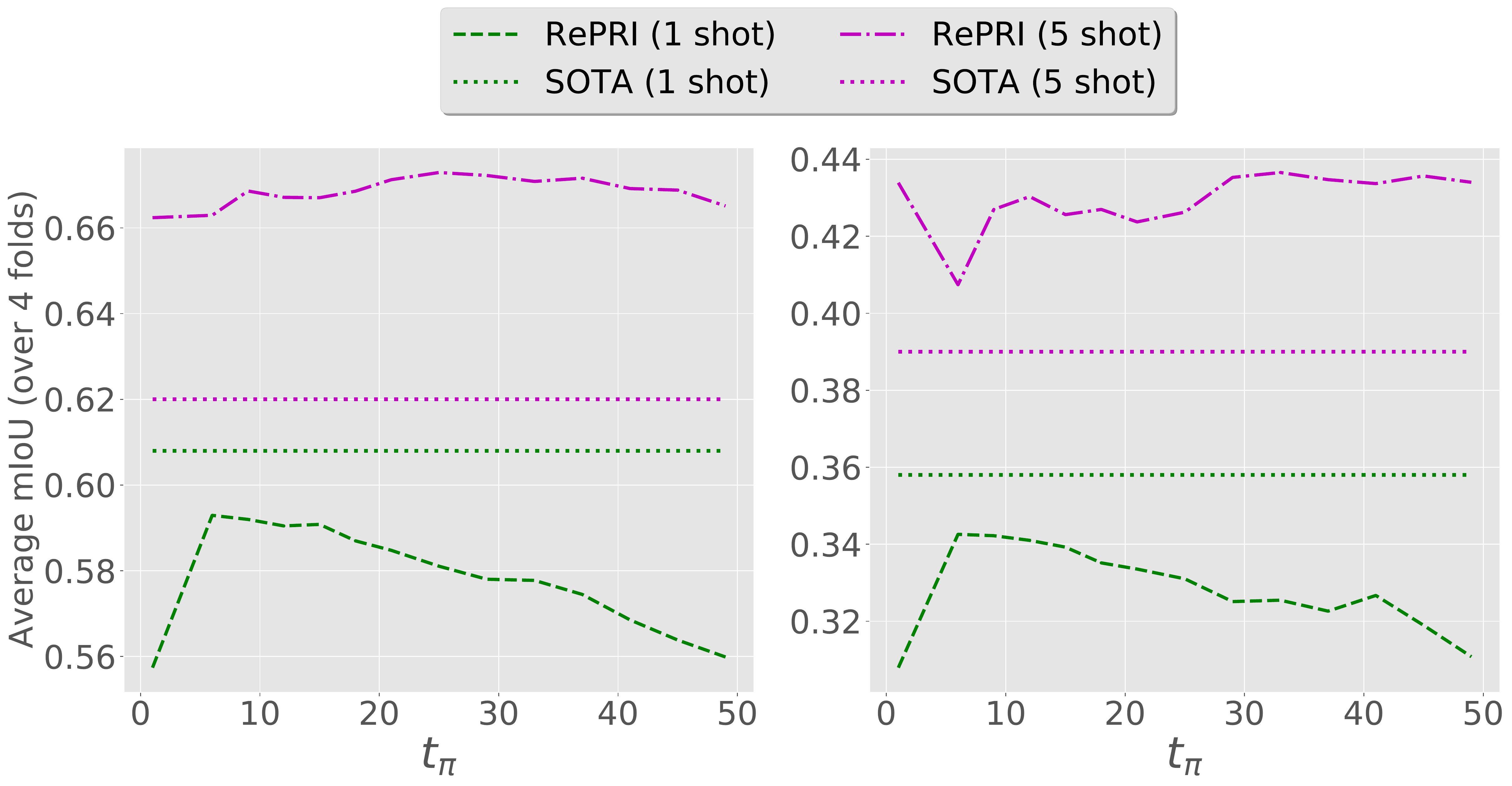}
            \caption{Average mIoU (over 4 folds) as a function of $t_\pi$ on PASCAL-$5^i$ (left) and COCO-$20^i$ (right) .}
            \label{fig:ablation_tpi}
        \end{figure}

        \paragraph{Influence of the parameter $t_\pi$} In Fig. \ref{fig:ablation_tpi}, we plot the averaged mIoU (over 4 folds) as a function of $t_\pi$ varying over the full range $t_\pi \in [1,50]$. For 5-shot, the performances are stable and remain largely above SOTA for all $t_\pi$. As for the 1-shot case, the range $[5,15]$ yields roughly similar results. While selection of optimal $t_\pi$ would lead to performance gains in each setting, in the paper, we used a single value of $t_\pi= 10$ for all the settings.

\begin{figure}%
        \centering
        \subfloat[ Relative error $\delta$ distribution of our current method, at initialization $\delta^{(0)}$ and after 10 gradient iterations $\delta^{(10)}$.]{{\includegraphics[width=.46\columnwidth]{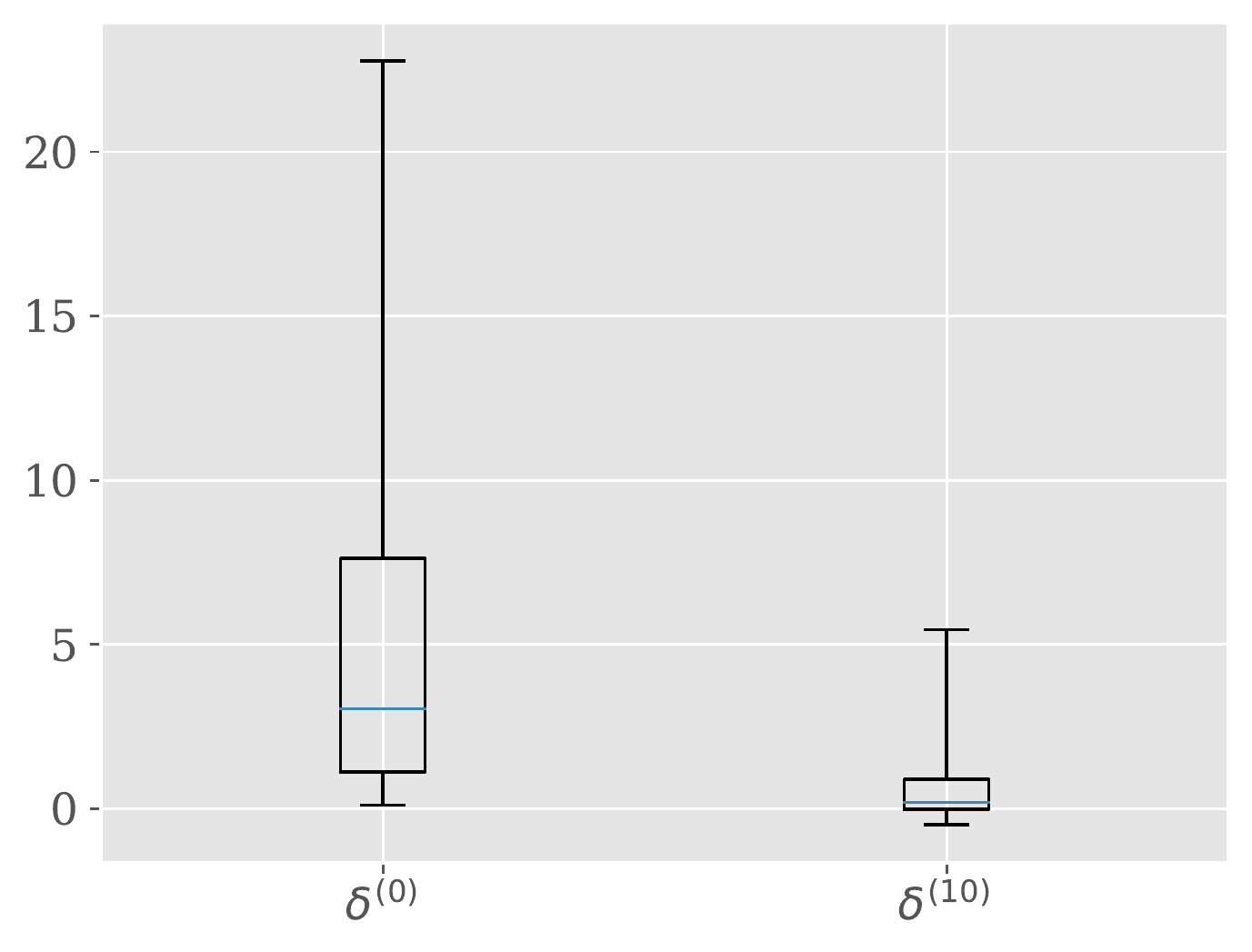} }\label{fig:delta_boxplot}}%
        \quad
        \subfloat[Mean-IoU versus enforced relative foreground size error $\delta$ in the parameter $\pi^{(0)}$.]{{\includegraphics[width=.46\columnwidth]{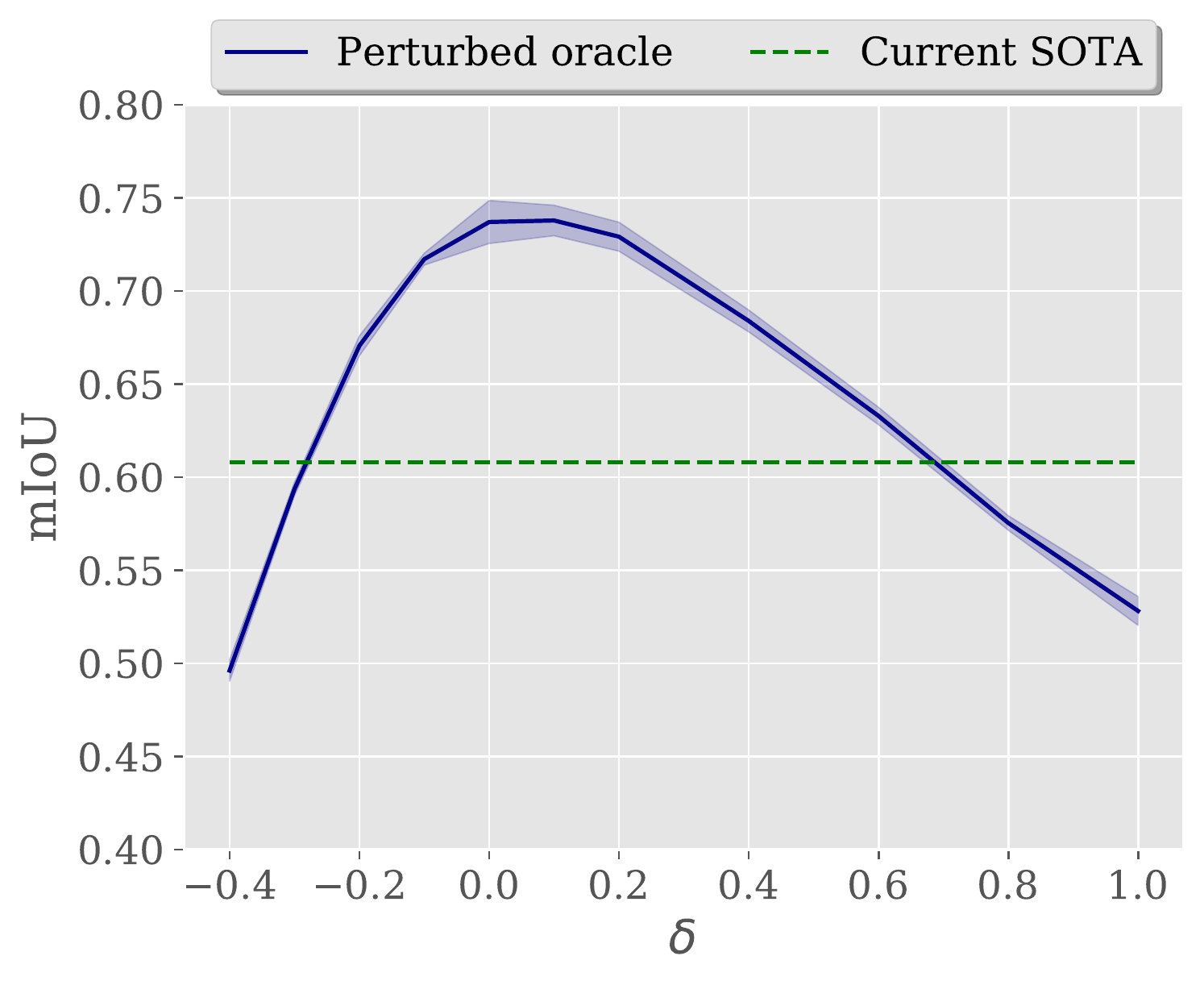} }\label{fig:oracle_perturbed}}
        \caption{Experiments on $\pi$ misestimation. Both figures are computed using 5 runs of 1000 1-shot tasks, each on the fold-0 of PASCAL$5^i$.}%%
    \end{figure}

        \vspace{-1em}
        \paragraph{Influence of parameter $\pi$ misestimation} Precisely knowing the foreground/background (B/F) proportion of the query object is unrealistic. To quantify the deviation from the exact B/F proportion $\pi^*$, we introduce the relative error on the foreground size:
        \begin{equation}
            \delta^{(t)} = \frac{\pi^{(t)}_1}{\pi^*_1} - 1,
        \end{equation}
        where $\pi^*_1$ represents the exact foreground proportion in the query image, extracted from its corresponding ground truth, and $\pi^{(t)}_1$ our estimate at iteration $t$, which is derived from the soft predicted segmentation.
        As observed from Fig. \ref{fig:loss_comparison}, the initial prototype often results in a blurred probability map, from which only a very coarse estimate of the query proportion can be inferred and used as $\pi^{(0)}$. The distribution of $\delta$ over 5000 tasks is presented in Fig. \ref{fig:delta_boxplot}. It clearly shows that the initial prediction typically provides an overestimate of the actual query foreground size, while  finetuning the classifier $\bm \theta$ for 10 iterations with our main loss (Eq. \ref{eq:main_objective}) already provides a strictly more accurate estimate, as conveyed by the right box plot in Fig. \ref{fig:delta_boxplot}, with an average $\delta$ around 0.7.

        Now, a natural question remains: \textbf{how good does the estimate need to be in order to approach the oracle results?} To answer this, we carry out a series of controlled experiments where, instead of computing $\pi^{(t)}$ with Eq. \eqref{eq:prior_init}, we use a $\delta$-perturbed oracle at initialization, such that $\pi^{(t)}_1 = \pi^*_1 (1 + \delta)$. Each point in Fig. \ref{fig:oracle_perturbed} represents the mIoU obtained over 5000 tasks for a given perturbation $\delta$. Fig.  \ref{fig:oracle_perturbed} reveals that exact B/F proportion is not required to significantly close the gap with the oracle. Specifically, foreground size estimates ranging from -10\% to +30\% with respect to the oracle proportion are sufficient to achieve 70\%+ of mIoU, which represents an improvement of 10\% over the current state-of-the art.
        This suggests that more refined size estimation methods may significantly increase the performance of the proposed method.

    \begin{table}[t]
            \centering
            \small
            \caption{Number of tasks performed per second, and the corresponding mIoU performances on PASCAL-5$^i$.}
            \begin{tabular}{lrrrr}
                \toprule
                  & \multicolumn{2}{c}{1-shot} & \multicolumn{2}{c}{5-shot} \\
                 \cmidrule(lr){2-3}\cmidrule(lr){4-5}
                 Method & FPS & mIoU & FPS & mIoU \\
                 \midrule
                 RPMMS \cite{rpmm} & 18.2 & 51.5 & 9.4 & 57.3 \\
                 PFENet \cite{pfenet} & 15.9 & 60.8 & 5.1 & 61.9 \\
                 RePRI (ours)  & 12.8 & 59.1 & 4.4 & 66.8 \\
                 \bottomrule
            \end{tabular}
            \label{tab:runtimes}
        \end{table}

    \paragraph{Computational efficiency} We now inspect the computational cost of the proposed model, and compare to recent existing methods. Unlike prior work, we solve an optimization problem at inference, which naturally slows down the inference process. However, in our case, only a single prototype vector $\w \in \mathbb R^C$, where we recall $C$ is the feature channel dimension, and a bias $b \in \mathbb R$ need to be optimized for each task.
    Furthermore, in our setting $C=512$, and therefore the problem can still be solved relatively efficiently, leading to reasonable inference times. In Table \ref{tab:runtimes}, we summarize the FPS rate at inference for our method, as well as for two competing approaches that only require a forward pass. We can observe that, unsurprisingly, our method reports lower FPS rates,
    without becoming unacceptably slower.
    % Nevertheless, despite the initial belief that our method would incur considerable computational costs,
    The reported values indicate that the differences in inference times are small compared to, for example, the approach in \cite{pfenet}. Particularly, in the 1-shot scenario, our method processes tasks 3 FPS slower than \cite{pfenet}, whereas this gap narrows down to 0.7 FPS in the 5-shot setting.

\section{Conclusion}
    Without resorting to the popular meta-learning paradigm, our proposed RePRI achieves new state-of-the-art results on standard 5-shot segmentation benchmarks, while being close to best performing approaches in the 1-shot setting. RePRI is modular and can, therefore, be used in conjunction with any feature extractor regardless how the base training was performed. Supported by the findings in this work, we believe that the relevance of the episodic training should be re-considered in the context of few-shot segmentation, and we provide a strong baseline to stimulate future research on this topic. Our results indicate that current state-of-the-art methods may have difficulty with more challenging settings, when dealing with domain shift or conducting inference on tasks whose structures are different from those seen in training---scenarios that have been overlooked in the literature. These findings align with recent observations in few shot classification \cite{chen2018closer,cao2019theoretical}. Furthermore, %as demonstrated in our results,
    embedding more accurate foreground-background proportion estimates appears to be a very promising way of constraining the inference, as demonstrated with the significantly improved results obtained by the oracle.
    Our implementation is publicly available online: \url{https://github.com/mboudiaf/RePRI-for-Few-Shot-Segmentation}.

% \clearpage
{%\small 
\bibliographystyle{ieee_fullname}
\bibliography{egbib}
}

\clearpage
\appendix
    \onecolumn

\section{Domain shift experiment}
    In Table \ref{tab:cross_domain_description_appendix}, we show the details of the cross-domain folds used for the domain-shift experiments. Also, in Table \ref{tab:COCO2PASCAL_results_appendix}, the per-fold results of the same experiment are available.

    \begin{table}[h]
        \centering
        \small
        \caption{Cross-domain folds.}
        \resizebox{\textwidth}{!}{\begin{tabular}{lccccc}
             & Dataset & Fold 0 & Fold 1 & Fold 2 & Fold 3 \\
             \toprule
             \multirow{7}{*}{\begin{tabular}{c} Excluded  \\ from \\ training \end{tabular}} & \multirow{7}{*}{MS-COCO} & Person, Airplane, Boat,    &     Bus, T. light, Bicycle,      &     Car, Fire H., Bird,    &  Motorcycle, Stop, Cat, \\
                                        &         & P. meter, Dog, Elephant   &     Bench, Horse, Bear,           &    Train, Sheep, Zebra     & Truck , Cow, Giraffe,  \\
                                        &         & Backpack, Suitcase, S. ball,   &     Umbrella, Frisbee, Kite,   &  Handbag, Skis, B. bat, & Tie, Snowboard, B.glove,   \\
                                        &         & Skateboard, W. glass, Spoon,   &     Surfboard, Cup, Bowl,   &  T. racket, Fork, Banana, & Bottle, Knife, Apple,    \\
                                        &         & Sandwich, Hot dog,   &     Orange, Pizza, Couch,   &  Boroccoli, Donut, P.plant, & Carrot, Cake, Bed,    \\
                                        &         & Chair, D. table, Mouse,   &     Toilet, Remote, Oven,  &  TV, Keyboard, Toaster,  & Laptop, Cellphone, Sink,    \\
                                        &         & Microwave, Fridge, Scissors   &     Book, Teddy &  Clock, Hairdrier  & Vase, Toothbrush    \\
             \midrule
             \multirow{2}{*}{\begin{tabular}{c} Test  \\ classes \end{tabular}} & \multirow{2}{*}{PASCAL-VOC} & Airplane, boat, chair, & Horse, Sofa, & Bird, Car, P.plant & Bottle, Cat, \\
             & &  D. table, Dog, Person & Bicycle, Bus & Sheep, Train, TV & Cow, Motorcycle \\
             \bottomrule
        \end{tabular}}
        \label{tab:cross_domain_description_appendix}
    \end{table}

    \begin{table}[h]
            \centering
            \small
            \caption{Per-fold domain-shift results on COCO-20$^i$ to PASCAL-5$^i$ experiment. Best results in bold.}
            \resizebox{\textwidth}{!}{
                \begin{tabular}{lcccccacccca}
                     & & \multicolumn{5}{c}{1 shot} & \multicolumn{5}{c}{5 shot} \\
                     \cmidrule(lr){3-7}\cmidrule(lr){8-12}
                     Method & Backbone & Fold-0 & Fold-1 & Fold-2 & Fold-3 & Mean & Fold-0 & Fold-1 & Fold-2 & Fold-3 & Mean \\
                     \toprule
                     RPMM \cite{rpmm} (ECCV'20)  & \multirow{3}{*}{ResNet50} & 36.3 & 55.0 & 52.5 & 54.6 & 49.6 & 40.2 & 58.0 & 55.2 & 61.8 & 53.8 \\
                     PFENet \cite{pfenet} & & 43.2 & \textbf{65.1} & \textbf{66.5} & 69.7 & 61.1 & 45.1 & 66.8 & 68.5 & \textbf{73.1} & 63.4 \\
                     RePRI (ours) & & \textbf{52.2} & 64.3 & 64.8 & \textbf{71.6} & \textbf{63.2} & \textbf{56.5} & \textbf{68.2} & \textbf{70.0} & \textbf{76.2} & \textbf{67.7} \\
                     \midrule
                     Oracle-RePRI & ResNet50 & 69.6 & 71.7 & 77.6 & 86.2 & 76.2 & 73.5 & 74.9 & 82.2 & 88.1 & 79.7 \\
                     \bottomrule
                \end{tabular}
            }
            \label{tab:COCO2PASCAL_results_appendix}
        \end{table}

\section{Results of the 10-shot experiments}

    In Table \ref{tab:10_shot_results_appendix}, we give the per-fold results  of the 10-shot experiments.

    \begin{table}[h]
    \centering
    \small
    \caption{Per-fold 10-shots results on PASCAL-5$^i$ and COCO-20$^i$. Best results in bold.}
    \resizebox{\textwidth}{!}{
        \begin{tabular}{lcccccacccca}
             & & \multicolumn{5}{c}{PASCAL-5$^i$} & \multicolumn{5}{c}{COCO-20$^i$} \\
             \cmidrule(lr){3-7}\cmidrule(lr){8-12}
             Method & Backbone & Fold-0 & Fold-1 & Fold-2 & Fold-3 & Mean & Fold-0 & Fold-1 & Fold-2 & Fold-3 & Mean \\
             \toprule
             RPMM \cite{rpmm} (ECCV'20)  & \multirow{3}{*}{ResNet50} & 56.1 & 68.2 & 53.9 & 52.3 & 57.6 & 30.9 & 39.2 & 28.2 & 34.0 & 33.1 \\
             PFENet \cite{pfenet} &  & 63.1 & 70.6 & 56.6 & 58.2 & 62.1 & 36.9 & 43.9 & 38.9 & 39.1 & 39.7 \\
             RePRI (ours) & & \textbf{65.7} & \textbf{71.9} & \textbf{73.3} & \textbf{61.2} & \textbf{68.1} & \textbf{41.6} & \textbf{48.2} & \textbf{42.1} & \textbf{44.5} & \textbf{44.1} \\
             \midrule
             Oracle-RePRI & ResNet50 & 75.6 & 81.0 & 82.1 & 75.6 & 78.6 & 57.5 & 64.7 & 56.6 & 56.1 & 58.7 \\
             \bottomrule
        \end{tabular}
    }
    \label{tab:10_shot_results_appendix}
    \end{table}

\clearpage

\section{Qualitative results}
\begin{figure}[h]
        \centering
        \includegraphics[width=\columnwidth]{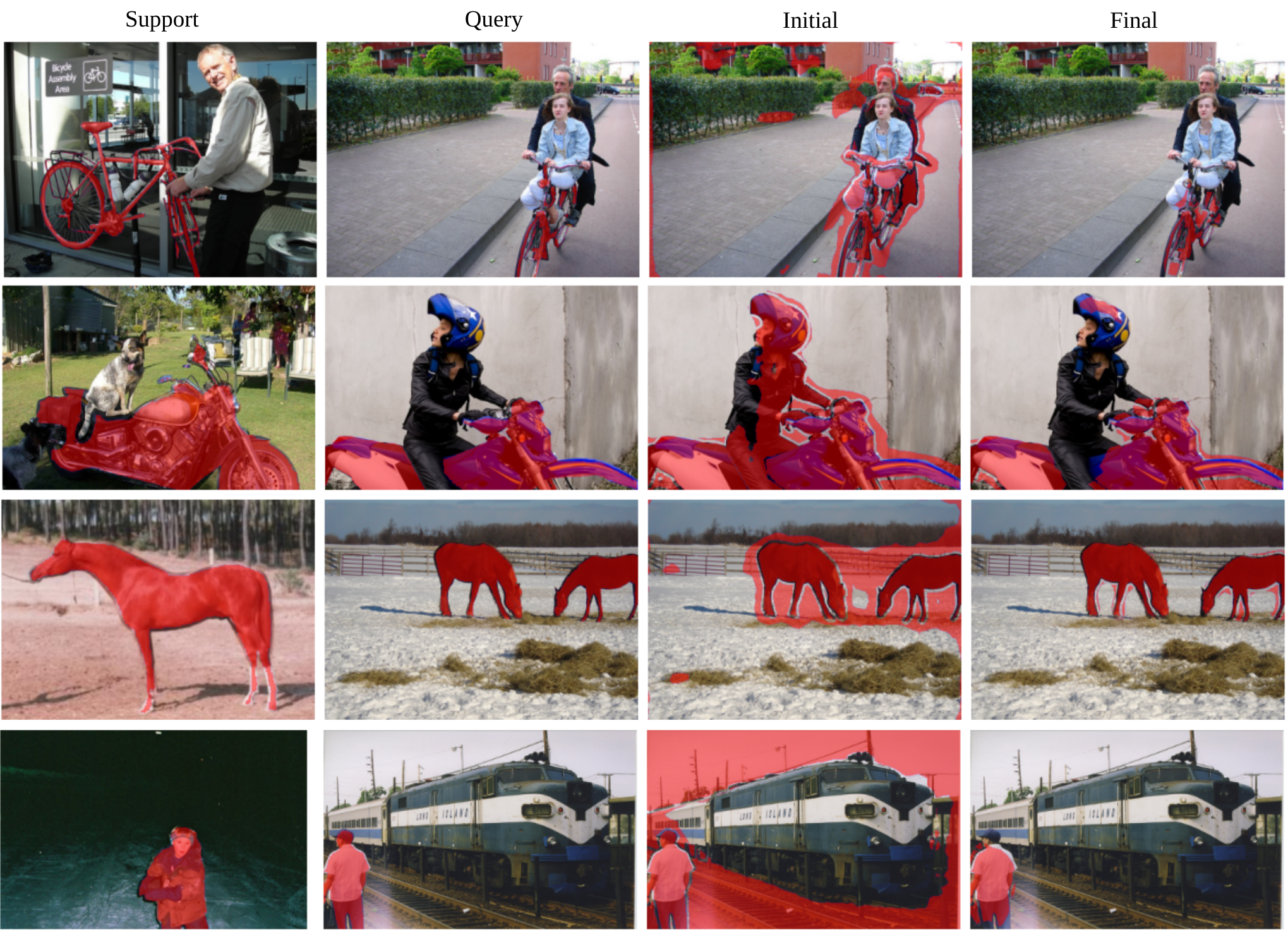}
        \caption{Qualitative results on PASCAL $5^i$. \textit{Initial} column refers to the predictions right after initializing the prototypes, while \textit{Final} column refers to the prediction after running our inference. Best viewed in colors in high resolution.}
        \label{fig:qualitative_results}
    \end{figure}

    In Figure \ref{fig:qualitative_results}, we provide some qualitative results on PASCAL-$5^i$ that show how our method helps refining the initial predictions of the classifier. 

\end{document}